# RePaint-Enhanced Conditional Diffusion Model for Parametric Engineering Designs under Performance and Parameter Constraints

Ke Wang, Nguyen Gia Hien Vu, Yifan Tang[1], Mostafa Rahmani Dehaghani, G. Gary Wang
Product Design and Optimization Laboratory (PDOL)
School of Mechatronic Systems Engineering
Simon Fraser University
Surrey, BC V3T 0A3

**Abstract**

*This paper presents a RePaint-enhanced framework that integrates a pre-trained performance-guided denoising diffusion probabilistic model (DDPM) for performance- and parameter-constraint engineering design generation. The proposed method enables the generation of missing design components based on a partial reference design while satisfying performance constraints, without retraining the underlying model. By applying mask-based resampling during inference process, RePaint allows efficient and controllable repainting of partial designs under both performance and parameter constraints, which is not supported by conventional DDPM-base methods. The framework is evaluated on two representative design problems, parametric ship hull design and airfoil design, demonstrating its ability to generate novel designs with expected performance based on a partial reference design. Results show that the method achieves accuracy comparable to or better than pre-trained models while enabling controlled novelty through fixing partial designs. Overall, the proposed approach provides an efficient, training-free solution for parameter-constraint-aware generative design in engineering applications.*
***Keywords***: *Generative Design, Diffusion Model, RePaint*

## 1 Introduction

Conceptual design creates solutions to fulfill predefined requirements and usually occurs at the beginning of the product design [1]. Effective execution is challenging because early-stage design is often subjective, qualitative, and uncertain [2]. The interdependence between engineering characteristics makes the process even further complex [2]. Designers seek creativity through the conventional paths such as brainstorm [3], checklist [4], lateral thinking [5], mind map [6]. However, those methods are often limited by the designers' knowledge bases and individual inspiration. As a result, finding the right novel conceptual design demands professional engineers with extensive experience and skills. However, the current pace of training new engineers is inadequate to meet the rapidly growing market demand [7]. According to an investigation by the Boston Consulting Group, the U.S. requires approximately 400,000 new engineering positions annually, but nearly one-third remain unfilled due to the shortage of qualified engineers. Furthermore, this talent gap is projected to persist until 2030 [7].

To address this challenge, two approaches have emerged: large-language-model-assisted design methods based on language representations and generative design approaches based on parametric representations. The first approach leverages large language models (LLMs) to assist designers in stimulating creativity and facilitating ideation during conceptual design. Designers can explore abstract concepts, obtain design briefs and discover relevant engineering principles through importing different prompts to the LLMs [8,9]. Some studies further explore the role of LLMs in the conceptual design process [10]. However, this approach only operates within the linguistic domain and focused more on supporting designers' tasks rather than producing concrete, parameterized design solutions. Their outputs often require further interpretation and translation by human designers before they can be efficiently realized in practice [9]. In addition, using LLMs to promote design generation exhibits difficulties in generating complex geometry and struggles to adapt to designs with a large number of parameters [11].

The second approach, generative design on parametric representations using artificial intelligence (AI) techniques, can directly create new parameterized designs under constraints via generative models [12]. This collaboration between designers and AI tools throughout the design process enables design paradigms that significantly expand the space of possible designs beyond what human designers alone can produce[13]. In addition, compared with text-based representations typically used in LLM outputs, parametric design representations explicitly capture the relationships between design features and engineering performance, thereby supporting key decision-making in complex engineering problems with multiple performance objectives [14]. As this work focuses on the generation of new designs, the second approach is studied in this study.

As a type of data-driven solution, generative models enable the efficient generation of new designs based on historical designs [15,16]. Their promising performance has been demonstrated in engineering design applications, including the design of airfoils [17,18], airplane shapes [19], layout patterns [20], wheels [21], and metamaterial [22]. Currently, multiple generative models have shown impressive performance in design generation, including Generative Adversarial Networks (GAN) [23], Variational Autoencoders (VAE) [24], and Denoising Diffusion Probabilistic Models (DDPM) [25,26]. Among these models, DDPM is a class of latent variable models inspired by considerations from non-equilibrium thermodynamics, which restore data from sampling noises via the denoising process [26]. It has demonstrated superior image synthesis capabilities and outperformed both GAN and VAE in terms of training performance and stability [27]. Overall, DDPM exhibits stable training behavior and flexible sampling characteristics, which make it a suitable choice for parametric design generation.

---

[1] Corresponding author: yifan_tang_2@sfu.ca.

Furthermore, the conditional DDPM incorporates additional information (such as part of an image or category label) as a generation condition to guide data synthesis compared to standard DDPM [28]. The conditional DDPM can utilize the performance indicators as its generation condition to guide design generation [29] and has already been applied in the design of battery structure [30], ship hulls [31], and topology optimization [32]. While conditional diffusion models have shown promising results in performance-guided design generation, a key limitation hinders their practical application in real-world parametric design scenarios. Once trained, a conditional DDPM typically requires retraining to accommodate new or modified generation constraints. In particular, introducing additional parameter constraints or generating designs based on partial designs cannot be readily handled by the pre-trained model alone, which limits its applicability in interactive and exploratory design scenarios.

To overcome this limitation, we propose a new framework that integrates the conditioning mechanism RePaint [33] with a performance-guided DDPM. By leveraging mask-based resampling during inference, the proposed approach enables the completion of full designs directly from partial designs while satisfying the target performance requirements, without retraining the underlying diffusion model. This design completion capability allows the generation process to flexibly incorporate arbitrary patterns of incomplete design. Therefore, the proposed approach enables designers to freely impose constraints on design parameters for design generation without retraining the DDPM each time the conditions are altered. In this work, the proposed method aims to achieve three objectives:

1. To enable fast conditional design generation without retraining for new conditions,
2. To complete incomplete designs under engineering performance and parameter constrains; and,
3. To generate novel designs based on partial designs.

The remaining paper is structured as follows. This paper reviews the generative design, DDPM, and RePaint in Section 2. The methodologies of the pre-trained model and the proposed framework are detailed in Section 3. All experimental results evaluating the performance of the proposed framework are summarized in Section 4. The discussion is presented in Section 5, followed by a conclusion in Section 6. The code and detailed results can be found at https://github.com/KeWang-atn/RePaint-cDDPM.git.

## 2 Related Works
### 2.1 Generative Design

In the last decade, rapid advancements in computing power and artificial intelligence (AI) technologies have enabled computers to generate conceptual and simulated solutions with expected performance using generative AI [34]. In this context, generative design is an approach that uses computational algorithms to generate and optimize solutions based on predefined design conditions and performance criteria [12]. Autodesk expands the definition of generative design to be "an advanced, algorithm-driven process, sometimes enabled by AI, used to explore a wide array of design possibilities that meet predefined criteria set by engineers or designers" [35]. Today, generative AI has become a key enabling technology for achieving generative design.

However, conventional generative design methods are typically tailored to specific problems and cannot generalize well. Additionally, this process requires extensive computational resources, particularly when multiple design parameters are involved. As a result, there is a need for a model that can handle diverse design constraints [36]. Fortunately, as discussed in Section 1, the rise of advanced generative models has shown promise in addressing these complex problems [31,37–40].

### 2.2 Denoising Diffusion Probabilistic Models

DDPM [26] is a powerful generative model. Its functionality is achieved through a forward process and a backward process. The forward process adds Gaussian noise to the original data over multiple steps until the data becomes indistinguishable from random noise. The core concept of the backward process is to learn the noise-adding behavior in the forward process and gradually remove the noise to reconstruct data that closely resembles the original distribution [26]. This approach leads to more stable training compared to other generative models, such as GAN, resulting in highly realistic and diverse outputs [27]. DDPM has demonstrated promising performance in image generation tasks and 3D model synthesis, producing state-of-the-art results with fewer artifacts and higher fidelity [38]. The simplicity and robustness of the diffusion process also make DDPM a powerful tool to quickly generate novel designs.

As discussed in Section 1, conditional DDPM improves the overall performance by incorporating additional information. The additional information can directly affect the model output and guide the data generation by correcting its errors [26,41]. In conditional DDPMs, performance-guided DDPMs [27] utilizes the guidance module to guide the data synthesis process instead of directly adding the extra label to condition the predicted noise. The guidance module is responsible for predicting the difference between the performance of the synthesized data in each denoising iteration and the target performance and making corrections accordingly [27]. Therefore, the performance-guided DDPM can adapt to different synthesis purposes by modifying the guidance module.

### 2.3 RePaint

RePaint is a conditioning mechanism, highly compatible with generative engineering designs tasks due to the ease of implementation and precise control over the generated data [33]. RePaint was initially proposed to solve the free-form image inpainting problem, which utilizes the iterative noise reduction capability of DDPMs to generate unknown regions based on the known ones [33]. In each denoising iteration of the diffusion model, RePaint combines the known part with the generated unknown part to produce a complete result and passes it to the next denoising iteration, as shown in Fig. 1 [32]. Furthermore, to prevent discontinuities between the generated and known regions,

RePaint employs a noise addition and resampling strategy, where noise is added to the known regions to match the noise level of the generated parts before being combined in each iteration. Then, each denoising iteration is repeated multiple times to integrate the two parts. RePaint is implemented through concatenation and can adapt to arbitrary shapes of incomplete input [33].

In the workflow of RePaint displayed in Fig. 1, the incomplete data or design is used as input for RePaint. The light shaded area represents the missing part of the input data. $X_{t-1\sim q}$ is the input data after aligning the noise level at the time step ($t$-1) via the forward process, $q$, of the diffusion model [26,41].

While $X_t$ represents the generated part that comes from the pre-trained diffusion model and is initially sampled from random Gaussian noise, and $X_{t-1\sim p_\theta}$ is generated data denoised by the reverse process, $p_\theta$, of the pre-trained diffusion model from the last iteration. Then, the white area in the mask has a value of 1 and can keep the data of $X$ in the corresponding area. On the contrary, the dark area shares value 0 and can remove data from the corresponding area. After extracting with masks, RePaint concatenates the known and generated parts as the final denoise data at time ($t$-1), $X_{t-1}$, and passes it as input to the next denoising iteration [33].

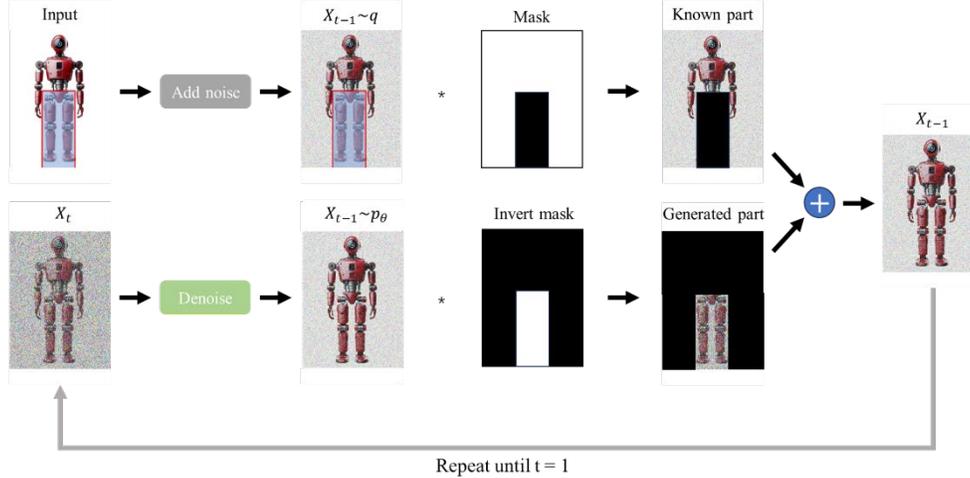

**Fig. 1 Data splicing method of RePaint [33]**

Due to the application of the mask in the inference process, RePaint can be directly applied to pre-trained models to adapt to arbitrary shapes of an incomplete image without retraining, and modifying the original model, while conventional conditional models require to learn the shape of the mask in the training, and fail to generalize to unseen masks [33]. Applying RePaint to the field of parametric design generation enables flexible conditioning by fixing design parameters via masks (e.g., fixing the length and height of a design with known values to generate the width or other parameters), and can also adapt to different subsets of known parameters (e.g., the known information changes from the height to the width or other parameters). Despite the advantages, the inference speed of RePaint is significantly slower than GAN, making it challenging to apply in systems that require fast response times [33].

## 3 Methodology

This section outlines the methodology for integrating the RePaint method into a diffusion model for engineering design applications.

### 3.1 Dataset

In this work, we test the proposed method in two datasets, ship hull design dataset and airfoil dataset. The ship hull design dataset, which is derived from Bagazinski et al. [31], comprises 82,168 parametric ship hull designs. This ship hull dataset also includes multiple performance indicators, such as volume, wave drag, etc. All the ship designs are represented using a data format proposed on Ship-D [39], which consists of 45 geometry-related parameters and could be found at https://github.com/KeWang-atn/RePaint-cDDPM.git.

In the parametric design vector of the ship hull design, the first parameter, the length overall ($LOA$), is the hull length in meters, while the remaining parameters expressed as ratios directly or indirectly related to $LOA$. Additionally, Ship-D provides a series of geometric formulas that enable the reconstruction of the 3D geometry of the hull using these 45 design parameters. All ship designs within the training dataset adhere to 49 algebraic constraints defined in Ship-D. These constraints must be satisfied to determine the feasibility of the generated ship hulls. A detailed documentation outlining these constraints for ship hull design parameters is available at [39]. Furthermore, in the study on ship hull design, we use the resistance coefficient, $C_t$, as the engineering performance indicator in the generation condition.

Besides the ship hull design dataset, the UIUC airfoil dataset [42] is adopted as another case study in this work to further assess the generality of the proposed method in different types of engineering design problems. The UIUC airfoil dataset contains a large collection of two-dimensional airfoil geometries with diverse shape characteristics, which have been widely used as benchmarks for aerodynamic analysis and data-driven design studies. Each airfoil is represented by a set of ordered surface coordinates describing its upper and lower profiles, enabling a compact yet expressive parametric representation. The lift-to-drag ratio coefficient, $C_l/C_d$, is provided with airfoil design and used as an engineering performance indicator in the generation condition in studies related to the airfoil dataset.

## 3.2 Conditional Denoising Diffusion Probabilistic Model

In the first step, a model that can generate designs according to the expected performance is required as a pre-trained model of the proposed framework. The diffusion model mentioned in Section 2.2 is a suitable tool. The original diffusion model was proposed to solve the problem of image generation [26]. However, this model is not directly applicable to the parameter-based engineering design generation. The introduction of C-ShipGen helps to eliminate this gap; it is a conditional tabular diffusion model [43] built for ship hull design generation [31]. Built upon ShipGen [40], a diffusion model for ship design, C-ShipGen was trained on the Ship-D dataset to generate new hull designs with specific volume and optimal drag resistance.

In this work, we aim to apply constraints on the engineering performance of the generated design rather than optimize it for specific performance indicators. Thus, based on the C-ShipGen model [31], we make some slight modifications and train a conditional model that generates design with a specific engineering performance value as the pre-trained model.

The training and sampling processes are detailed in ALGORITHM 1 and ALGORITHM 2, respectively.

**ALGORITHM 1:** Training algorithm for the conditional diffusion model

| | |
|---|---|
| 1: | **Input:** |
| 2: | Training dataset $D = \{(x_0^i, C^i)\}_{i=1}^N$ |
| 3: | conditional noise prediction network $\epsilon_\theta(\cdot)$ |
| 4: | **repeat** |
| 5: | $\quad (X_0, C) \sim D$ |
| 6: | $\quad X_0 \sim q(X_0)$      % extract hull designs |
| 7: | $\quad t \sim Uniform(\{1, \ldots, T\})$    % randomly pick a time step |
| 8: | $\quad \epsilon \sim N(0, \mathbf{I})$      % sample noise |
| 9: | $\quad$ Take a gradient descent step on: |
| | $\quad \nabla_\theta \|\epsilon - \epsilon_\theta(\sqrt{\bar{\alpha}^t}x_0 + \sqrt{1-\bar{\alpha}^t}\epsilon, t, C)\|^2$   % model training |
| 10: | **until** converged |

The training algorithm for the conditional diffusion model is summarized in ALGORITHM 1. Lines 2 and 3 show the first step, loading the prediction network $\epsilon_\theta(\cdot)$ and the training dataset $D$, consisting of $N$ ground truth designs $x_0$ and condition label $C$. The design $x_0$ includes multiple design parameters (e.g., length overall, and design draft) which control the geometry feature of the design, while the condition label is composed of performance indicators (e.g., resistance coefficient, and lift-to-drag ratio coefficient) and the environmental variables (e.g., velocity, and temperature) that are contributing to performance calculation. Then, during the training procedure of the condition model (lines 4 to 10), the data and label are extracted from the training set as the first step. Secondly, some noises are added to the design data by noise adding operator $q(\cdot)$ to establish the learnable dataset (line 6). After that, a random time step $t$ and noise $\epsilon$ are sampled (line 7 and 8). Subsequently, the prediction network $\epsilon_\theta$ predict the noise based on the noised design data, time step and condition, where $\bar{\alpha}^t$ denotes the cumulative product of preservation coefficients up to step $t$, determining the proportion of the original data retained after $t$ diffusion steps. And then the prediction network is trained based on the difference between the predicted noise and the sample random noise (line 9). The training procedure (line 4 to 9) is repeated until the difference between the predicted noise and sample random noise converges. Furthermore, the predictor network used in this work consists of a 6-layer Residual neural network with 512-dimensional latent features, 128-dimensional time and condition embedding.

**ALGORITHM 2:** Conditional DDPM sampling process with performance guidance (Revised from [31])

| | |
|---|---|
| 1: | **input** $C$      %load generation condition |
| 2: | $X_t \sim N(0, I)$    % sample random noise as the start points |
| 3: | **for** $t = T, \ldots, 1$ **do**    % small noise added after denoising |
| 4: | $\quad Z \sim N(0, I)$ if $t > 1$, else $z = 0$   % denoise with guidance |
| 5: | $\quad X_{t-1} = \frac{1}{\sqrt{\alpha_t}}\left(X_t - \frac{1-\alpha_t}{\sqrt{1-\bar{\alpha}_t}}\epsilon_\theta(X_t, t, C)\right) + \sigma_t(Z(1-\gamma)) +$ $\gamma \nabla_{X_t} f_\phi(y\|X_t) - \lambda \nabla_{X_t}\left(C_t - P_{perf}(X_t)\right)^2$ |
| 6: | **end for** |
| 7: | **return** $X_0$ |

The sampling process with performance guidance adopted in conditional DDPM is summarized in ALGORITHM 2. In line 1, the generation condition label $C$ is loaded as input, where this label should include the expected engineering performance indicators and corresponding environmental variables. Then, in line 2, a random noise is sampled as the start point $X_t$ of the following procedure. After that, lines 3 to 6 demonstrate the denoise and guidance process. The starting point $X_t$ remove the noise predicted by the prediction network and receive the guidance from the gradient produced by classifier $f_\phi$ and performance predictor $P_{perf}$ to generate the denoised design data $X_{t-1}$. Then, the denoise and guidance process is repeated by $T$ times to generate the final output design, $X_0$. Moreover, the $Z$ refers to a small noise added to the denoised design data, $\sigma_t$ controls the amount of the added noise, $\alpha_t$ refers to the preservation coefficients, $\gamma$ and $\lambda$ are the coefficients control the contribution of the classifier and performance predictor. The classifier and predictor used in this work are adopted from the work of C-ShipGen [31] with detailed architecture information shown in Table **1** [30].

**Table 1 Architecture of the feasibility classifier and the resistance coefficients predictor [31]**

| | Classifier, $f_\phi$ | Predictor, $P_{C_t}$ |
|---|---|---|
| Type of model | Multi-layer perceptron (MLP) | Residual neural network (ResNet) |
| Layer number | 5 | 6 |
| Latent feature# | [128, 64, 64, 64] | [512, 512, 512, 512, 512] |
| Output feature# | 1 | 1 |
| Activate function | SiLU, Sigmoid | SiLU |
| Loss function | Binary cross-entropy | Mean squared error |

## 3.3 Integrating RePaint with Pre-trained Model

As discussed in Section 2.3, Repaint helps to impose a strong constraint on the parameter values of the design. This section aims to introduce RePaint into the pre-trained conditional DDPM by modifying the inference algorithm, and the proposed framework is showed in Fig. 2. The detailed implementation of this model is described in ALGORITHM 3.

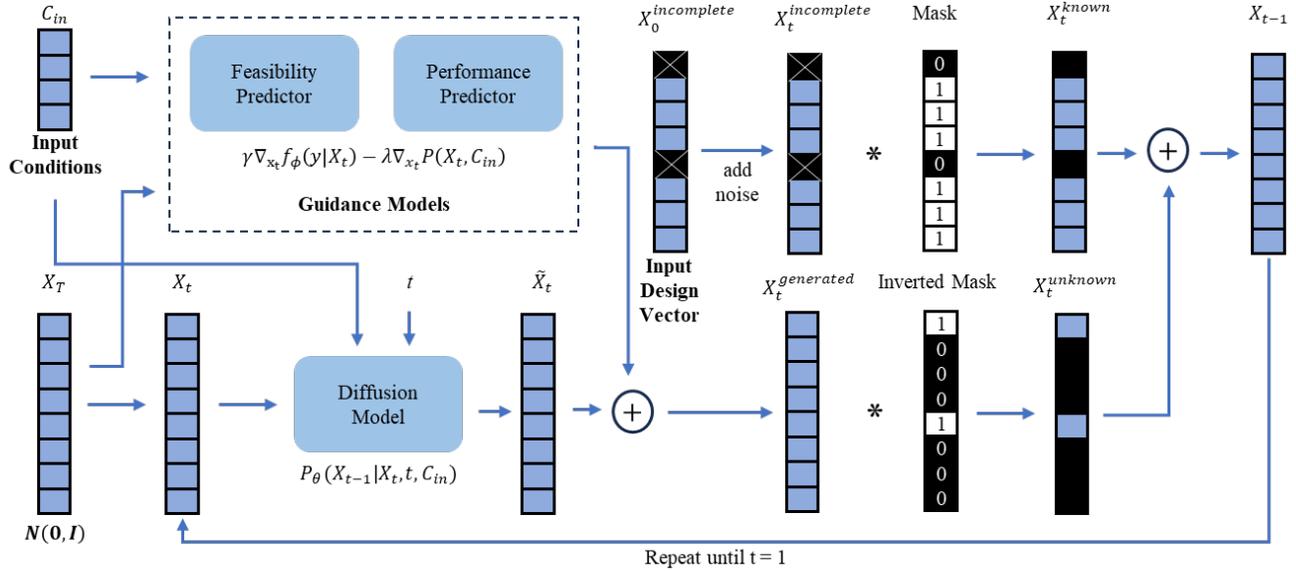

**Fig. 2 The overview of the proposed model**

In ALGORITHM 3, this work utilizes RePaint to make a pretrained diffusion model to synthesize an engineering design with partially known design parameters. This method combines the unknown parameters $X^{generated}$ and the known design parameters $X^{known}$, where $X^{generated}$ is sampled from the pre-trained model via the reverse process and $X^{known}$ is sampled through the forward process by a mask $m$ and an inverse mask with a Boolean value in each iteration (line 9), as shown in Fig. 2. The combined design parameters are used as the denoised result $X_{t-1}$ and passed to the next denoising iteration as input, eventually to produce the complete design $X_0$ (line 9). This process repeats inside each denoising iteration of DDPMs twenty times to eliminate the influence caused by data splicing [33] (line 4), and the $U$ refers to the resampling iteration number.

**ALGORITHM 3:** The sampling process after adding the RePaint method to DDPM

1:   **input** $C$
2:   $X_t \sim N(0, I)$      % sample random noise as input
3:   **for** $t = T, \ldots, 1$ **do**      % denoising iteration
4:     **for** $u = 1, \ldots, U$ **do**      % resampling iteration
5:       $\epsilon \sim N(0, \mathbf{I})$ if t >1, else $\epsilon = 0$      % sample noise for noise alignment
6:       $X_{t-1}^{known} = \sqrt{\overline{\alpha_t}} X_0 + (1 - \overline{\alpha_t}) \epsilon$      % noise alignment
7:       $z \sim N(0, \mathbf{I})$ if $t > 1$, else $z = 0$      % small noise added after denoise
8:       $X_{t-1}^{generated} = \frac{1}{\sqrt{\alpha_t}} \left( X_t - \frac{1-\alpha_t}{\sqrt{1-\overline{\alpha_t}}} \epsilon_\theta(X_t, t, C) \right) + \sigma_t(Z(1-\gamma)) + \gamma \nabla_{x_t} f_\phi(y|X_t) - \lambda \nabla_{x_t} \left( C_t - P_{perf}(X_t) \right)^2$
9:       $X_{t-1} = (1-m) \odot X_{t-1}^{generated} + m \odot X_{t-1}^{known}$      % data splicing
10:     **if** $u < U$ and $t > 1$ **then**      % stop resampling in last denoising iteration
11:       $X_t \sim N(\sqrt{1-\beta_{t-1}} X_{t-1}, \beta_{t-1} \mathbf{I})$      % restore noise level for the next resampling
12:     **end if**
13:   **end for**
14:   **end for**
15:   **return** $X_0$

### 3.4 Performance Evaluation

In this section, we test the proposed method in terms of accuracy and diversity among both ship hull design and airfoil design.

#### 3.4.1 Accuracy.

To evaluate the accuracy of RePaint's output designs' performance, we implement the mean absolute percentage error (MAPE) between the expected performance value and the generated designs' performance value as the metric. The lower MAPE value indicates that the generated design's performance is closer to the expected value and more accurate. Moreover, due to differences in the representation of design between the ship hull design dataset and the airfoil design dataset, we design different experiments for both case studies.

In the case of ship hull design, resistance coefficient ($C_t$) is the expected performance indicator. Then, we design two tasks: single and multiple known parameter tasks.

In the single known parameter task, each hull design parameter is individually fixed from the reference design, and new ship hull designs are generated. Furthermore, the proposed method is expected to generate new ship hull designs with expected resistance coefficient, $C_t$, under the condition of

RePaint.

In the multiple known parameter task, subsets of parameters are fixed simultaneously. Two operations for generating subsets are studied: (1) incrementally fixing parameters from 6 to 43 to assess the effect of increasing the fixed parameter count, and (2) fixing each hull component group to generate new ship hull designs. The design parameters are grouped by hull regions to balance computational cost and experimental relevance: midship cross section (Parameters 6–9), bow geometry (Parameters 10–18), stern geometry (Parameters 19–29), and bulb geometry (Parameters 30–43). More details of parameter numbers and corresponding physical definitions are referred to [39].

The generation accuracy of proposed method is measured using the mean absolute percentage error (MAPE) of the generated ship hull designs' resistance coefficient ($C_t$) compared to the expected $C_t$ value. We generate 512 designs for evaluation in all cases introduced beyond. For consistency, the generation condition and reference design input of both tasks are set to the same (length overall: 333, Beam: 42.624, Draft: 11.28, Depth: 29.064, resistance coefficient: 0.0003, Froude number: 0.43). Details of the result can be found in Sections 4.2.1 and Section 4.2.2.

For the airfoil design case, we randomly select 100 unseen airfoil designs as reference design. Then, through the RePaint condition mechanism, fix part of the airfoil to generate 100 new airfoil designs for each reference airfoil design. The fixing part is increased from one-eighth to seven-eighths of the entire airfoil in steps of one-eighth. All the airfoil generation is conducted under the condition of reference designs' lift-to-drag ratio coefficient value. Hence, the performance of the proposed method in this case is measured by MAPE of the lift-to-drag ratio coefficient $C_l/C_d$. The results are displaced in Section 4.2.3.

### 3.4.2 Diversity.

After evaluating the accuracy and stability of the proposed method, we also assess its ability to generate diverse and novel designs. In this section, Precision and Recall for Distributions (PRD) and Maximum Mean Discrepancy (MMD) are calculated for both generated ship hull design and airfoil design. PRD reflects the trade-off between sample quality and diversity, whereas MMD measures the overall distributional discrepancy between generated and reference samples. For PRD, we use the method introduced in Sajjadi et al. [44] to calculate the precision and recall of generated designs in both case studies. After that, the result is presented in the form of a PRD curve, which jointly characterizes sample quality and coverage by visualizing the trade-off between the precision and recall across different operating points [44]. This enables a more comprehensive assessment of generative performance than single-value metric. For MMD, we apply a Gaussian radial basis function (RBF) kernel to compute the mean embedding difference between generated parts and the training set. It measures how similar the generated and actual sample distributions are. A smaller MMD value indicates a closer distributional match, whereas a larger MMD value suggests that the generated designs deviate further from the training distribution and exhibit greater novelty. The detailed results can be found in Section 4.2.4.

## 4 Results

To ensure the effectiveness of the model, we first trained two diffusion models on both ship hull and airfoil dataset (Section 4.1). To assess the accuracy of RePaint in engineering design generation, we first analyzed the generation accuracy of the proposed method in Section 4.2.1, 4.2.2, and 4.2.3. Then, we measure the diversity in both two study cases in Section 4.2.4. All the training and inference in this work were conducted on a platform with NVIDIA RTX 6000 pro Blackwell GPU.

### 4.1 Model Training

Before applying the RePaint conditional mechanism, a pre-trained model needs to be prepared. Accordingly, two performance-guided DDPM models are trained on the ship hull and airfoil datasets using the method introduced in Section 3.2. Notably, in the ship hull design case, the dataset provides the feasibility labels and judgment standard for the training feasibility classifier, but the airfoil design case does not offer this. Hence, there is no classifier in the conditional model trained on the airfoil dataset. After finishing model training, we record the training time and measure inference times with and without applying the RePaint condition mechanism to generate 512 designs. The generated designs are used to measure the difference from the expected value and calculate the performance MAPE. As mentioned in Section 3.1, the performance indicator used in airfoil generation is the lift-to-drag ratio coefficient $C_l/C_d$ while the ship hull generation uses the resistance coefficient $C_t$ as performance indicator to constrain the design generation. Details on the time cost, data used for training, and performance MAPE of the generated designs without applying RePaint are in Table 2. Patrial ship hull designs generated with RePaint are displayed in Fig. 3, which shows the reference ship hull design on the left side and generated ship hull design on the right side.

**Table 2 Time cost, data used and performance of pre-training model**

| Dataset | Training | Inference without RePaint | Inference with RePaint | Training Data Size | Performance MAPE |
|---------|----------|---------------------------|------------------------|--------------------|------------------|
| Airfoil | 6h 25min | 4.072s | 83.721s | 38802 | 5.314% |
| Ship hull | 7h 45min | 3.199s | 64.173s | 82168 | 5.736% |

As shown in Table 2, the pre-trained models for both the airfoil and ship hull datasets require more than 6 hours of training and achieve a performance error of approximately 5.5% MAPE. During inference, the incorporation of the RePaint mechanism increases the computation time from a few seconds to over one minute due to the resampling operations introduced in Section 3.3. Nevertheless, the inference time with RePaint remains significantly shorter than the overall training time. Moreover, the proposed method enables the pre-trained models to generate new designs under additional constraints on the design parameters. In other words, the proposed framework allows well-trained models to accommodate new design constraints without requiring model retraining. This capability highlights the advantage of the proposed approach in enabling constraint-aware design generation in a training-free manner. More details about the

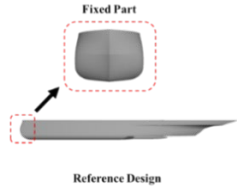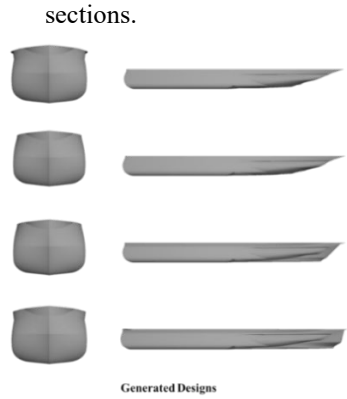

**Fig. 3 Generating new hull designs while keeping the same bow structure**

### 4.2 Performance evaluation
#### 4.2.1 Fixing Individual Parameters of Ship Hull

In this subsection, we evaluate the accuracy of individually fixing each of the 44 ship hull design parameters through the single known parameter task introduced in Section 3.4.1. Among all the results, the result of fixing an individual parameter of the midship cross-section (Parameters 6 to 9: Beam at Chain, Deadrise Angle, Radius of Chine and Radius of Keel) is selected as the representation and shown in Table 3. The results of fixing each of the rest of the individual parameters are omitted here for brevity.

**Table 3 $C_t$ MAPE of fixing individual parameters (Parameters 6 to 9)**

| Fixed-Parameter | Drag Coefficient Mean | $C_t$ MAPE (%) |
|---|---|---|
| 6 | 0.000284 | 5.780 |
| 7 | 0.000326 | 8.045 |
| 8 | 0.000280 | 7.297 |
| 9 | 0.000276 | 8.986 |

Among the results shown in Table 3, the case of fixing Parameter 6 has the lowest MAPE for $C_t$, 5.780%, which is close to the performance of designs generated by the pre-trained model (5.736%). After that, the other three cases show the $C_t$ MAPEs, which are slightly higher than the case of fixing Parameter 6, Beam at Chain, but still stay around 8%. Fixing single parameters via RePaint can provide users with designs with performance close to that of the pre-trained model.

#### 4.2.2 Fixing Multiple Parameters of Ship Hull

In addition to fixing parameters individually, the new framework allows multiple ship hull design parameters to be fixed simultaneously as a condition and does not limit the number and selection of the design parameters, which offers more flexibility. In this section, we evaluate the effect of both the number and selection of fixed parameters on model performance by the multiple known parameter tasks mentioned in Section 3.4.1. The result of increasing the number of fixed parameters is shown in Table 4. The result of more combinations of fixed parameters is omitted for brevity.

Table 4 summarizes the performance of the proposed model when the number of fixed ship hull design parameters increases from 2 to 38. Among those results, the generated designs have a $C_t$ MAPE value that fluctuates around 6%, except for fixing Parameters 6 to 18. In this case, the $C_t$ MAPE value suddenly rises to 20.157%. However, continuing to include more fixed parameters does not bring a worse result, but makes the result stay about 6.5%, as shown in the cases that fix Parameters 6 to 29 and Parameters 6 to 36. Moreover, the case fixing Parameters 6 to 43 has an outstanding result, 6.391% MAPE, even though it fixes the most parameters. This phenomenon suggests that variations in the fixed regions lead to corresponding changes in the generated regions. The fixed sections of the reference design may act as generation conditions that are not fully captured by the model under the current generation space and expected performance constraints. Further investigation into the underlying causes is left for future work.

**Table 4 Result of fixing different subsets of parameters**

| Fixed-Parameter | Drag Coefficient Mean | $C_t$ MAPE (%) |
|---|---|---|
| 6-7 | 0.000323 | 7.256 |
| 6-8 | 0.000316 | 5.924 |
| 6-9 | 0.000289 | 4.774 |
| 6-18 | 0.000250 | 20.157 |
| 6-29 | 0.000321 | 6.523 |
| 6-36 | 0.000321 | 6.477 |
| 6-43 | 0.000321 | 6.391 |

The effect of fixing parameters that belong to the same ship component is further investigated. Parameters 1 to 5 (Length of Bow Taper, Length of Stern Taper, Beam of Midship Deck, Depth of Hill, Beam at Stern Deck, and Design Draft) were not considered in this experiment since they control the main dimensions of the hull. Table 5 displays the result of fixing parameters for the hull components.

Table 5 summarizes the generation performance under configurations with one or two fixed components. Among the individual component cases, the case of fixing the stern geometry shows an abnormal performance bias of 18.657% $C_t$ MAPE. The increase in this error can also be due to changes in the efficient generation space, as mentioned in the analysis of fixing different numbers of parameters. Furthermore, according to the results of cases that fix multiple components, the influence of fixing the stern can be mitigated by combining with other components (Bow and Midship) to reduce $C_t$ MAPE to below 18.657%. Besides those cases, other cases that fix components can still produce designs with $C_t$ MAPE below 8%. Bow is another important geometry part on the other side of the ship hull and fixing it can generate designs with 6.749% in performance error. However, fixing it with other components brings a deterioration in the performance accuracy.

**Table 5 Results of different fixation schemes by ship components: midship (6-9), bow (10-18), stern (19-29), and bulb (30-43)**

| Fixed Components | Drag Coefficient Mean | $C_t$ MAPE (%) |
|---|---|---|
| Bow | 0.000319 | 6.749 |
| Stern | 0.000369 | 18.657 |
| Bulb | 0.000284 | 6.114 |
| Midship, Stern | 0.000331 | 9.286 |
| Bow, Stern | 0.000356 | 15.694 |
| Midship, Bulb | 0.000288 | 4.840 |
| Stern, Bulb | 0.00037 | 18.887 |
| Bow, Bulb | 0.00032 | 7.213 |

In conclusion, fixing multiple parameters of ship hull design simultaneously can still provide cases with good performance, but the selection of the components has a significant effect on the performance of the output design generated by RePaint due to the interactions between selected items. Some studies about potential influencing factors could be found in Appendix (Section A 0).

*4.2.3 Generation accuracy in airfoil design generation*

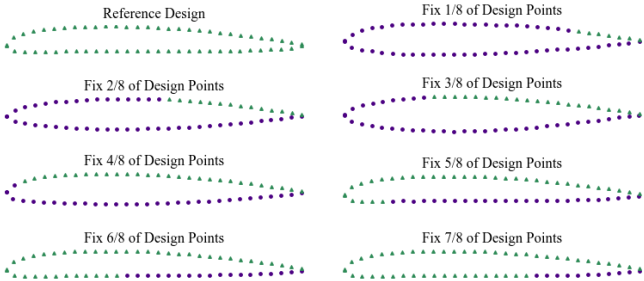

**Fig. 4 Generated airfoils with different proportions of fixed design points (green triangles: fixed points; purple dots: generated points)**

In this subsection, according to the experiments introduced in Section 3.4.1, we selected 100 reference airfoil designs and generated 100 new designs for each by applying the RePaint mechanism to retain a portion of the reference design. We repeat the experiment with different retention rates from one-eighth to seven-eighths as airfoil designs shown in Fig. 4. Then, we calculate the MAPE value between the lift-to-drag ratio coefficient value of generated designs and expected value as the generated design's performance error. The lower MAPE value indicating higher generation accuracy. The results are displayed in Table 6.

Table 6 reports the mean absolute percentage error (MAPE) of the lift-to-drag ratio ($C_l/C_d$) evaluated on the generated airfoil designs under different fixed-part ratios. For reference, the pre-trained conditional model without partial geometry constraints achieves a baseline MAPE of 5.314%, which represents the generation accuracy when the entire airfoil geometry is generated solely under performance conditioning. When only a small portion of the airfoil geometry is fixed (from the first 1/8 to the first 3/8), the generation error ranges from 5.67% to 6.60%, which is comparable to or slightly worse than the baseline performance of the pre-trained conditional model. A notable improvement is observed once more than half of the airfoil geometry is fixed. At a fixed ratio of 5/8, the MAPE decreases significantly to 3.80%, representing a substantial improvement over the baseline error of 5.314%. Further increasing the fixed portion continues to enhance performance alignment, with the lowest MAPE of 3.46% achieved when 7/8 of the geometry is fixed. These results demonstrate that while partial geometry constraints provide limited benefit at low fixed ratios, incorporating a sufficiently large portion of reference geometry enables the RePaint-based framework to outperform the pre-trained conditional model in terms of generation fidelity. This highlights the effectiveness of the proposed method for constraint-aware airfoil design generation and partial design completion without retraining the underlying model.

**Table 6 performance error of generated airfoil in different proportions of fixed design points**

| Fixed area | $C_l/C_d$ MAPE (%) |
|---|---|
| First 1/8 points | 5.669 |
| First 2/8 points | 6.023 |
| First 3/8 points | 6.604 |
| First 4/8 points | 5.865 |
| First 5/8 points | 3.799 |
| First 6/8 points | 3.598 |
| First 7/8 points | 3.458 |

*4.2.4 Diversity*

Generating diverse designs is another fundamental objective of generative design. To evaluate the diversity of the proposed method, we adopt the approach described in Section 3.4.2 to compute the precision and recall of the generated designs for both ship hull and airfoil cases. The resulting Precision and Recall are visualized as PRD curves in Fig. 5 and Fig. 6. respectively. In addition, the Maximum Mean Discrepancy (MMD) is calculated to quantitatively measure the distributional distance between the generated designs and the training data, with the results reported in Table 7.

For the ship hull design case, the diversity analysis focuses on designs generated by fixing a single design parameter using the RePaint-based conditioning mechanism. The corresponding precision, recall, and MMD values are evaluated, with the PRD curves presented in Fig. 5 and the MMD results summarized in Table 7, along with those from the airfoil case. We restrict the main analysis to the single-parameter fixing scenario for clarity, as the cases involving fixed components exhibit similar trends in both PRD and MMD metrics. The detailed results for component-level fixing are therefore provided in the Appendix (Section A 0) for completeness.

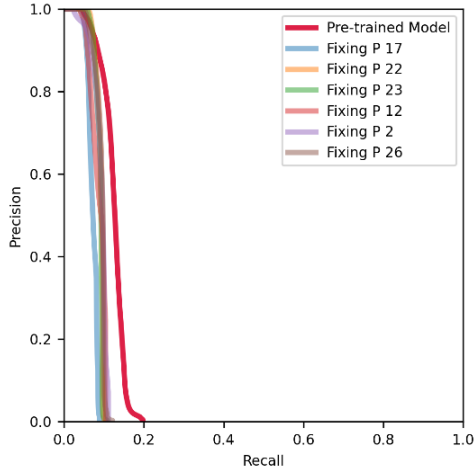

**Fig. 5 PRD curve of fixing single parameter of ship hull design**

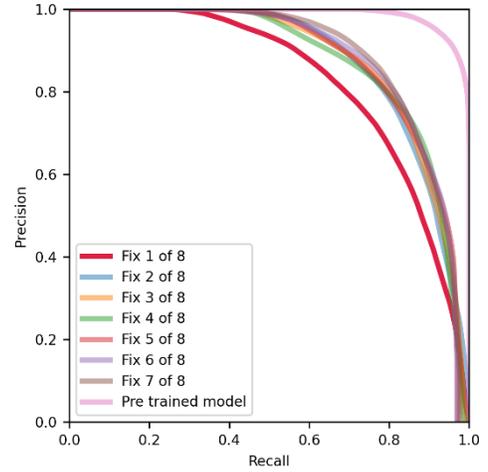

**Fig. 6 PRD curve of generated airfoil with different proportions of fixed design points**

Fig. 5 and Fig. 6 present the PRD curves of the generated designs under different partial-fixing settings for the two study cases. In Fig. 5, the PRD curves corresponding to the single-parameter fixing cases lies marginally to the left of that of the pre-trained conditional model, exhibiting similar precision–recall trends across the selected representative parameters. Additionally, the use of a single generation condition restricts the design coverage of the generated samples, resulting in recall values below 1 in this case study. In Fig. 6, a consistent shift of the PRD curves of the generated airfoil design can be observed as the partial design points are fixed. Compared with the pre-trained model, the PRD curves of the RePaint-based generation cases are slightly shifted inward in the precision–recall space, while maintaining high precision over a broad recall range. Across both figures, the PRD results indicate that the generated designs remain concentrated in regions of high precision, with only a moderate reduction in recall relative to the pre-trained conditional model.

**Table 7 MMD of generated designs in two study cases**

| Study case | Fixed area | MMD |
|---|---|---|
| Airfoil | First 1/8 | 0.0250 |
|  | First 2/8 | 0.0296 |
|  | First 3/8 | 0.0327 |
|  | First 4/8 | 0.0374 |
|  | First 5/8 | 0.0461 |
|  | First 6/8 | 0.0461 |
|  | First 7/8 | 0.0465 |
|  | No fixed part | 0.0138 |
| Ship hull | Parameter 2 | 0.220 |
|  | Parameter 12 | 0.215 |
|  | Parameter 17 | 0.252 |
|  | Parameter 22 | 0.221 |
|  | Parameter 23 | 0.220 |
|  | Parameter 26 | 0.330 |
|  | No fixed part | 0.097 |

**Table 7** reports the Maximum Mean Discrepancy (MMD) between the generated designs and the training data for both airfoil and ship hull study cases under different fixation settings. For the ship hull study case, the no-fix condition yields an MMD value of 0.097. When individual design parameters are fixed, the MMD values increase and vary across parameters, ranging from 0.215 to 0.330. Most fixed-parameter cases exhibit MMD values around 0.22 to 0.25, while Parameter 26 shows a comparatively higher MMD value of 0.330. For the airfoil study case, the MMD value for the no-fix condition is 0.0138. As the proportion of fixed design area increases from the first 1/8 to the first 7/8, the MMD increases progressively from 0.0250 to 0.0465. The MMD values show a monotonic rising trend with increasing fixed area and stabilize at approximately 0.046 once more than half of the design area is fixed. Overall, in both study cases, fixing design variables leads to higher MMD values compared to the no-fixed condition, with the airfoil case showing a gradual increase associated with the proportion of fixed area and the ship hull case showing parameter-dependent variability.

## 5   Discussion

The following subsections elaborate in-depth on the research findings presented in the results. Section 5.1 discusses the overall performance of RePaint in engineering design generation. Section 5.2 outlines some limitations awaiting improvement and discusses possible future directions.

### 5.1   Flexible Conditional Design Generation

This work demonstrates that the proposed RePaint-enhanced framework enables constraint-aware design generation by incorporating additional conditions into pre-trained conditional models without requiring retraining. Across both ship hull and airfoil design generation tasks, the method preserves high design quality while enabling controlled novelty through partial design fixing. The results shown in Section 4.2 indicate that RePaint is particularly effective for partial design completion and localized design modification. For ship hull design, fixing individual parameters generally maintains generation accuracy comparable to that of the pre-trained model, whereas fixing multiple parameters or entire components leads to performance variations that depend strongly on parameter interactions. In contrast, the airfoil design case exhibits a clearer trend, where fixing a sufficiently large portion of the geometry improves generation accuracy beyond the baseline model.

For the diversity analysis, despite the differences in design representations between the two case studies, Fig. 5 and Fig. 6 exhibit a consistent trend in their PRD behavior. In both cases, the PRD curves of designs generated using the RePaint-enhanced partial generation framework are shifted slightly inward relative to those of the pre-trained conditional model, indicating a modest reduction in distributional coverage. Importantly, this shift is limited in magnitude, and high precision is largely preserved, suggesting that the proposed RePaint-based approach remains capable of generating high-quality designs. This observation is further supported by the MMD results, which show an increase in distributional distance between the generated designs and the training data when partial constraints are imposed. In both study cases, the MMD values are higher for RePaint-based generation than for the no-fixed condition, confirming that the generated designs deviate from the original training distribution. However, the increase in MMD is moderate and remains stable across different fixation schemes, indicating a controlled and bounded distributional shift rather than excessive divergence. Taken together, the PRD and MMD analyses demonstrate that the observed reduction in coverage arises from the fixation of a subset of design parameters during generation, which reduces the effective design space explored by the model. This behavior does not reflect a loss of generative capability of the underlying model; instead, it indicates that the intrinsic diversity learned by the pre-trained model is preserved within a reduced design domain defined by the imposed partial constraints. As a result, the proposed framework can generate diverse and high-quality designs while explicitly incorporating user-specified design parameter conditions

From a computational perspective, RePaint provides a significant efficiency advantage as discussed in Section 4.1. Training the conditional models requires several hours (6–8 hours), whereas the RePaint condition mechanism adds extra constraints on design parameters without retraining and makes the conditional design generation within one to two minutes, despite the additional resampling overhead. This large reduction in computation time enables rapid incorporation of new design constraints while maintaining competitive performance accuracy (MAPE ≈ 5–6%), making the proposed framework well suited for interactive and iterative engineering design workflows.

### 5.2 Limitations

Despite the advantage that RePaint enables the incorporation of additional conditions into a pre-trained model without requiring retraining, several limitations remain. In design problems where strong interdependencies exist among design parameters, such as ship hull design, the generation accuracy is sensitive to the choice of which design parameters or components are fixed. Different selections of fixed regions can lead to noticeably different performance outcomes. This sensitivity suggests that the impact of geometric constraints is not uniform across the design space and depends on the underlying inter-relationships among parameters. The interaction between constraint selection, parameter coupling, and generation accuracy exhibits highly complex behavior, which is difficult to characterize analytically within the current framework. A systematic investigation into constraint-aware strategies that account for parameter dependencies is therefore left for future work.

## 6 Conclusion

This paper proposes a RePaint-enhanced conditional generative framework for performance- and parameter-constraint engineering design generation without retraining requirement. By applying a mask-based resampling strategy during inference, the proposed method allows selected design parameters or components to be fixed while preserving the generative capability of the original conditional model. Experiments on ship hull and airfoil design datasets demonstrate that the proposed framework can generate high-quality designs under diverse constraint configurations. While the effectiveness of constraint imposition depends on parameter interactions in strongly coupled design spaces such as ship hull design, the method shows clear advantages in partial airfoil design completion tasks. Diversity analysis confirms that RePaint preserves design quality while enabling controlled novelty through focused exploration of reduced-dimensional design subspaces. Moreover, the proposed approach offers substantial computational savings, reducing the cost of incorporating new constraints from hours of retraining to minutes of inference, making it well suited for interactive and iterative engineering design workflows.

**Acknowledgements**

Funding from the Natural Science and Engineering Research Council (NSERC) of Canada under the project RGPIN2019-06601 is gratefully acknowledged.

# Appendix
## A1. Accuracy of Predictor

To evaluate the accuracy of the predictor used in this paper, we conduct an experiment in ship hull case. We sample 100,000 ship hull designs as a testing set to predict their resistance coefficient, $C_t$, value by the predictor and calculate the $R^2$ and MAPE of the $C_t$. The comparison plot is shown in Fig. A1, the distribution of the $C_t$ value is also plot in Fig. A2 to jointly analyze the performance of the predictor.

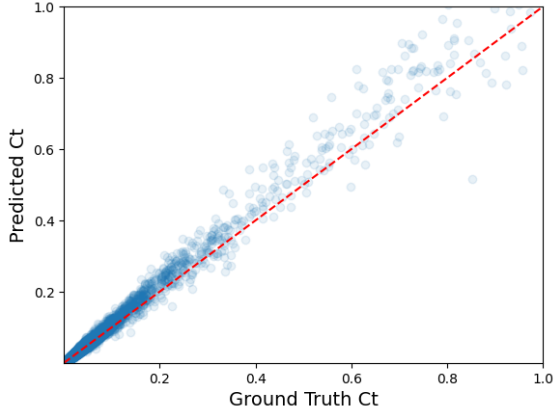
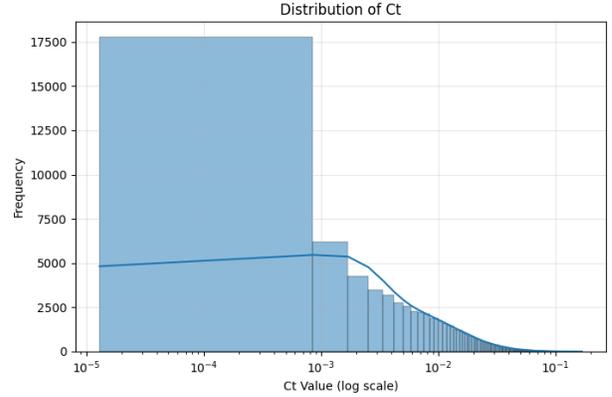

**Fig. A1 The reliability plot of the $C_t$ predictor used in this work**

**Fig. A2 The $C_t$ value distribution of the training set**

The predictor's MAPE $C_t$ and $R^2$ are 6.613% and 0.981. Fig. A1 demonstrates the overall performance of the predictor, the predicted $C_t$ tend to spread from the ground truth value as the $C_t$ value increases but still shows a clear linear relationship. Fig. A2 illustrates that the training set's $C_t$ value concentrates on the lower side, with an area less than 0.001, and its frequency gradually decreases with the increase of the $C_t$ value. Hence, we think the performance degradation of the predictor is caused by the nonuniform distribution of the training set for the $C_t$ value. In brief, the predictor exhibits good accuracy for small $C_t$ values and has the potential to address its shortcomings by optimizing its training set in the future.

## A2. Stability

To evaluate the stability of the proposed method's output, we conduct an experiment in ship hull design case. Firstly, we use a random ship hull design as a reference and repeat the design generation 30 times with fixed conditions, including the fixed parameters, components, and generation conditions, $C$, for data collection. Then, calculate the standard deviation of the resistance coefficient $C_t$ MAPE and feasibility ratio of generated designs for each case. A lower standard deviation indicates greater stability. By using the method introduced beyond, we first sample 512 designs using the pre-trained model 30 times without applying RePaint and obtain a mean of around 5.603 and a standard deviation of 0.180 for the $C_t$ MAPE across all trials. Moreover, the pre-trained model's generated designs were 100% feasible across all trials, yielding 512 designs that met the parameter-value constraints. Then, we repeat experiments on the case of fixing a single parameter and the case of fixing a single component. Their results are displayed in Table A1.

**Table A1 Standard deviation of $C_t$ MAPE and the rate of feasibility**

| Fixed Component | Std. of Ct MAPE | Std. of Feasibility |
|---|---|---|
| Mid. Cross Sec. | 0.058 | 0.000 |
| Bow | 0.172 | 0.000 |
| Stern | 0.122 | 0.003 |
| Bulb | 0.123 | 0.000 |
| Fixed Parameter | Std. of Ct MAPE | Std. of Feasibility |
| 6 | 0.144 | 0.000 |
| 10 | 0.117 | 0.000 |
| 19 | 0.192 | 0.000 |
| 30 | 0.218 | 0.000 |
| None RePaint | Std. of Ct MAPE | Std. of Feasibility |
| Pre-trained Model | 0.180 | 0.000 |

From Table A1, all of the standard deviation of $C_t$ MAPE for cases fixing a single component has a small value (smaller than 0.22), and the case that fixes the Midship Cross Section (Parameter 6 to 9) shows the most stable $C_t$ error, 0.058. In the field of feasibility rate, all instances can continue to generate over 500 designs (512 total) that meet the parameter-value constraints and maintain a near-zero standard deviation. When fixing single parameters, the first design parameters of each component are picked as the representative.

Even though the standard deviation of cases of fixing a single parameter is slightly higher than that of fixing a single component, they are kept at a small value and show a stable error. Meanwhile, the feasibility rate of fixing a single can also remain close to 100% with a slight standard deviation. More details can be found in https://github.com/KeWang-atn/RePaint-cDDPM.git. Therefore, the proposed method can maintain stable conditions for both $C_t$ error and feasibility rate.

**A3. Factors Affecting the Performance of RePaint**

In the ship hull design case, we find that the generation accuracy of the proposed method is affected by the selection of the fixed parameter or section as results shown in Section 4.2.2. Therefore, we further investigate the factors influencing generation accuracy in the ship hull design case. We propose three hypotheses that the performance of the generated design is influenced by (1) the distribution of parameter values in the training set, (2) the correlation between design parameters and the resistance coefficient ($C_t$), and (3) the pre-trained model under the same generation condition. The order of the tabular data could be another important influencing factor on the model; however, it directly influences the pre-trained model's learned representation and then indirectly affects conditional generation behavior after partial fixing. In this Section, we aim to study the factors that affect the application of RePaint. The order of parameters is not considered in this work and will be explored in future studies.

**A3.1 Effect of the Training Set.**

Since DDPM aims to generate data with a distribution close to the training set [26] to assess how data distribution in the training set affects design generation performance, we analyze MAPE of $C_t$ when fixing each parameter at its most and least frequent values. Each parameter's feasible range is divided into 20 intervals to determine the frequency distribution. The intervals with the highest and lowest frequencies are identified, and their midpoints are selected as representative values. By fixing each parameter at these values individually, new designs are generated and evaluated by MAPE of $C_t$. A significant MAPE difference between high- and low-frequency settings supports the first hypothesis. Otherwise, the first hypothesis cannot be established. We selected the results for Parameters 6 to 9 (Beam at Chain, Deadrise Angle, Radius of Chine and Radius of Keel) to discuss.

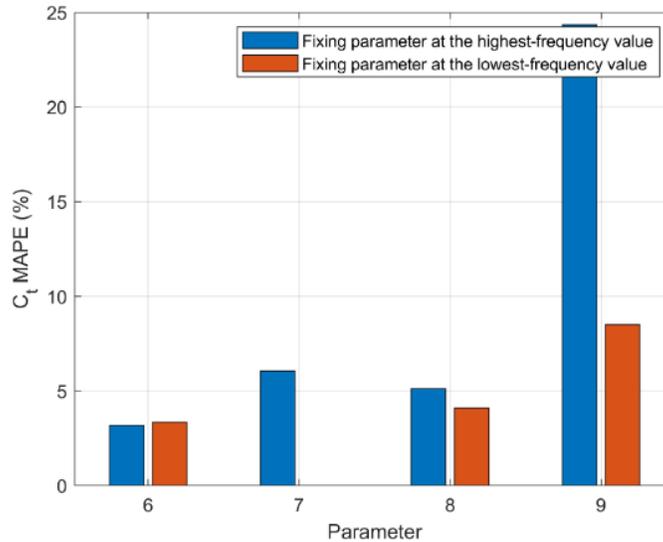

**Fig. A3** $C_t$ MAPE difference of fixing parameters 6 to 9 individually

In the results shown in Fig. A3, the only significant difference between the cases of fixing parameters at the highest frequency value and the lowest frequency value is in the results of Parameter 7. Fixing Parameter 7 at the highest frequency value can generate designs with $C_t$ MAPE around 6%, while no feasible design could be obtained when fixing the parameter at the lowest frequency value, followed by the case of fixing Parameter 9, around a 15% $C_t$ MAPE difference. In the results of fixing Parameters 6 and 8, the difference in $C_t$ MAPE is insignificant. Specifically, in the case of fixing Parameter 6, the results of fixing the parameter at the two extreme frequency values are almost the same. There seems to be no strong correlation between the frequency of parameter values and the performance of the final generated designs. The difference in the frequency of occurrence of fixed values generally does not directly lead to significant differences in the performance of the generated design. The first hypothesis that the performance of the generated design depends on the distribution of parameter values in the training set, does not hold.

**A3.2 Effect of Correlation Between Design Parameters and Performance.**

To evaluate the impact of parameter-performance correlation on framework performance, we calculate the total resistance coefficient ($C_t$) for all hull designs in the training set using the predictor from C-shipGen [31], based on draft and Froud number, $F_n$ conditions from Section 3.4.1. We then compute the correlation between each parameter and $C_t$ and compare these correlations with the $C_t$ MAPE observed when fixing individual parameters. A stronger correlation corresponding to higher $C_t$ MAPE would support the second hypothesis. Otherwise, the second hypothesis is not supported.

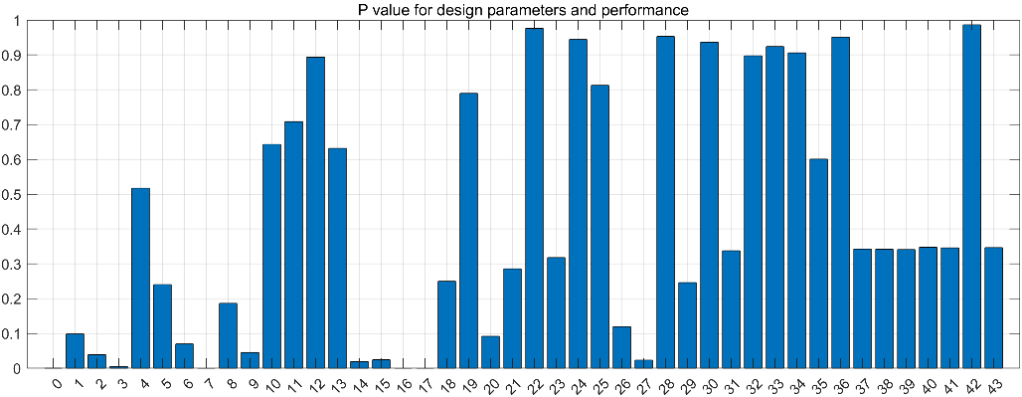

**Fig. A4 Correlation between design parameters and total resistance coefficient of the ship hull**

In Fig. A4, only ten parameters exhibit statistically significant correlations with $C_t$ (p-value < 0.05), including Parameters 0 (Length of Bow Taper), 2 (Beam at Midship Deck), 3 (Depth of Hull), 7 (Deadrise angle), 9 (Radius of Keel), 14, 15 (Constants to define curve for the midship width), 16, 17 (Constants for the curve that defines drift angle along z-axis), and 27 (Beam at Transom Chine). Comparing the result in Section 4.2.1 with the result of fixing a single parameter in Table 3, it is clear that similar $C_t$ MAPEs could be found in parameters with different correlations to $C_t$, such as the cases for Parameters 7 (Deadrise angle) and 8 (Radius of Chine). Therefore, we cannot conclude that the correlation with performance has a significant effect on the performance via RePaint, and the second hypothesis that the correlation between design parameters and total resistance coefficient of the ship affects the performance of the generated design is not supported.

**A3.3 Effect of the Pre-trained Model.**

RePaint is only a conditioning mechanism that guides the pre-trained model's inference. To test whether the pre-trained model influences the final output, we generate 512 design variations using only the pre-trained model, bypassing RePaint. We identify each parameter's most frequent value (assigning boundary values where applicable). These values are then fixed by RePaint to generate new designs, and their MAPEs of $C_t$ are evaluated. We study the performance of fixing other values in the value range. For efficiency, we divide the value range of the design parameters into ten equal parts and fix the value at each dividing point (a total of 11 points) to generate new designs for performance evaluation. If fixing high-frequency values from the pre-trained model yields a low MAPE, and at low frequencies a high MAPE, the third hypothesis is supported. Detailed results can be seen in Fig. A5.

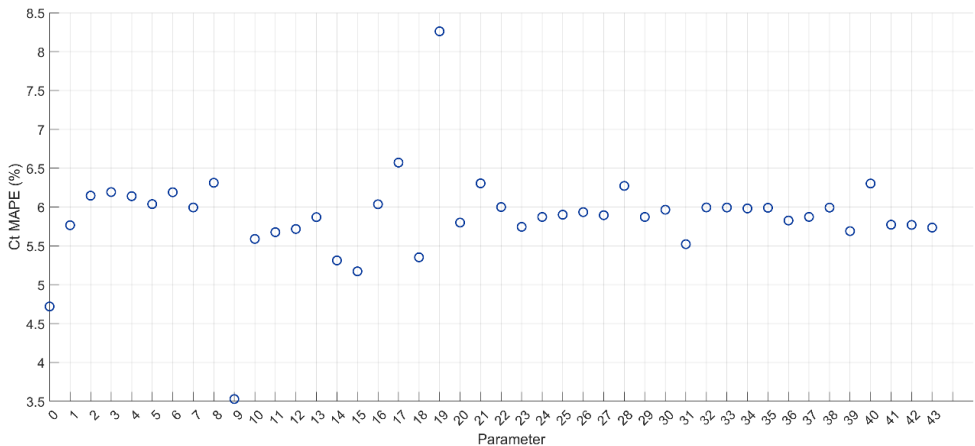

**Fig. A5 Ct MAPE based on the most frequent parameter value in the output of the pre-trained model**

In Fig. A5, the MAPE of $C_t$ based on the most frequent parameter value in the output of the pre-trained model is stable at around 6%. The highest MAPE could be found in the case of Parameter 19 (Lower stern taper bit), which reached around 8.2%, but is still acceptable. Afterward, in the case of Parameter 9 (Radius of Keel), its MAPE dropped to 3.5%, and the designs generated in this case showed a close performance with the generation condition. Overall performance in this study is close to the pre-trained model. RePaint can also fix the high-frequency value in designs generated by the pre-trained model to generate designs with similar performance. After that, we continue to further study the relationship between the value distribution and generated designs' performance. The results of Parameters 6 to 9 are representative and shown in Fig. A6 and Fig. A7.

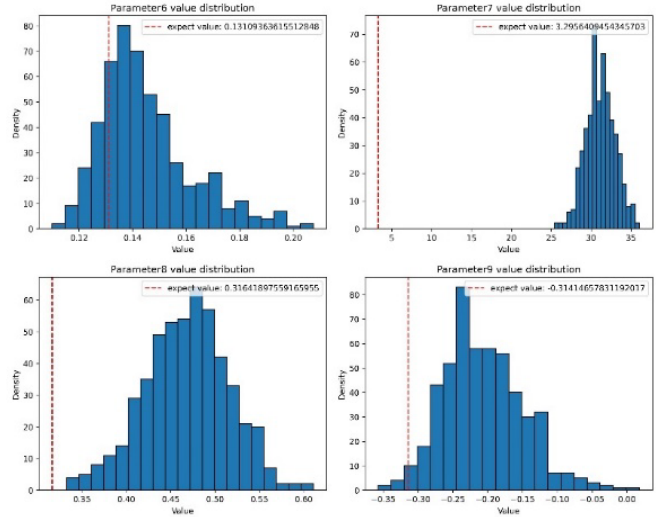

**Fig. A6 Distribution of the pre-trained model's output parameter values for Parameters 6 to 9**

Fig. A6 displays the distribution of values for the pre-trained model's output designs. The red dotted line represents the value in the reference design mentioned in the preceding sections. The values of the four parameters show non-uniform distribution and do not spread out the entire value range. For example, the value range of Parameter 6 (Beam at Chine) is from 0 to 0.6, but the distribution shown in Parameter 6 only exists from 0 to 0.21 and is blank in other areas. This phenomenon also occurs in the other three parameters. It illustrates that the pre-trained model tends to generate values in a relatively concentrated range and barely generates feasible designs in the other areas.

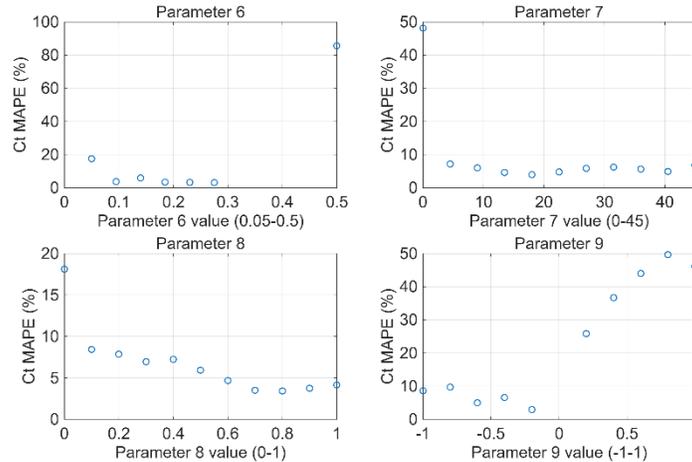

**Fig. A7 The Ct MAPE when fixing Parameters 6 to 9 at different values**

Results shown in Fig. A7 is the $C_t$ MAPE of fixing different values in the value ranges. Incomplete values are shown for Parameters 6 and 9, as no feasible designs were generated in those cases. Referring to **Error! Reference source not found.**, fixing the value located in the pre-trained model's output range can generate feasible designs with low $C_t$ MAPE. When the fixed value is outside the range, the performance of the generated design starts to become unstable. The generated designs' $C_t$ MAPE in Parameters 7 (Deadrise angle) and 8 (Radius of Chine) can be kept below 10% across the entire range, except in the area close to the low end. At the same time, it suddenly jumps to a high MAPE value, and the model even fails to generate feasible designs in Parameters 6 (Beam at Chine) and 9 (Radius of Keel) when the value rises over the pre-train model's output value range. Through the application of RePaint, fixing a value inside the pre-trained model's output range can ensure generating designs with a low $C_t$ MAPE, and fixing value outside the range can also help to explore different design spaces from that of the pre-trained model. Therefore, the third hypothesis receives a certain level of support.

**A4 Diversity of fixing different components of ship hull design**

By using the method introduced in Section 3.4.2, we calculate the precision and recall of designs generated by fixing components of ship hull design and present the result in form of PRD curve and shown in Fig. A8. Beside that, the MMD values are also calculated and displayed in Table A2.

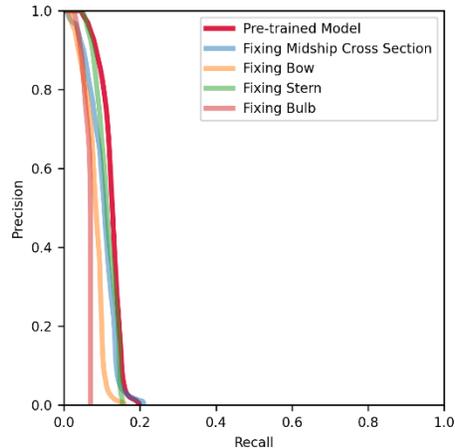

**Fig. A8 PRD curve of fixing a single component of ship hull**

Fig. A8 demonstrates the PRD results for the case of fixing a single design component. Similar to the case of fixing single-parameter, the PRD curves corresponding to fixing individual components consistently lie below that of the pre-trained conditional model. This shift indicates a moderate reduction in recall, suggesting that imposing component-level geometric constraints restricts the accessible design space to a subset of the original distribution. Nevertheless, the precision of the generated designs remains comparable to that of the pre-trained model, indicating that the proposed framework continues to produce high-quality designs despite the reduced distributional coverage. These results suggest that fixing a single component guides the generative process toward more constrained yet structurally meaningful design variations, while still preserving a reasonable level of diversity.

**Table A2 MMD of fixing a single component**

| Fixed Component | MMD |
| --- | --- |
| Mid. Cross Sec. | 0.296 |
| Bow | 0.404 |
| Stern | 0.340 |
| Bulb | 0.216 |
| None (pre-trained model) | 0.0969 |

The MMD values for fixing single components are reported in Table A2. The MMD values mainly fall within the range of 0.2 to 0.41, which is higher than those of the pre-trained model, which is 0.0969. This indicates that the generated designs exhibit greater distributional diversity. When different hull components are fixed, the increase in MMD suggests that the model explores broader variations in the unfixed regions, leading to a richer and more novel set of design outcomes. Fixing the Bow and Stern encourages the generator to produce more diverse global geometries. In contrast, fixing the Bulb results in the smallest increase. It can be attributed to its weaker cooperation with the global hull geometry. In this region, geometric variations exert a smaller influence on the overall hull form. Overall, these results indicate that the method can flexibly adjust its design generation process, maintain diversity while adapt to different fixed-component or fixed-parameter conditions.

**A5  Hyperparameters sensitivity.**
To assess the influence of hyperparameters on model performance, we examined the sensitivity of the resampling count $U$, the feasibility classifier guidance weight $\gamma$, and the performance predictor guidance weight $\lambda$. We only conduct this experiment in ship hull design case since the airfoil design case does not provide feasibility labels and corresponding judgment standard to training the classifier. Resampling is applied at every denoising step to mitigate discontinuities between the generated and reference parts. Following Lugmayr et al. [33], a resampling count of 20 is commonly recommended for practical applications; here, we tested $U = 20$ and $U = 30$ to evaluate their impact. The parameters $\gamma$ and $\lambda$ were each set to two levels (0.3 and 0.7), and a 2³ Full Factorial Sensitivity Design of Experiments (DOE) on $U$, $\gamma$ and $\lambda$ was followed. The mean absolute percentage error (MAPE) of the total resistance coefficient $C_t$ was then computed to quantify performance under each configuration. Finally, a three-way analysis of variance (ANOVA) was conducted to statistically evaluate the main and interaction effects of $\lambda$, $\gamma$, and $U$ on the $C_t$ MAPE. The results can be found in Table A3. To guide the fine-tuning of the hyperparameter, the interaction of two most significant factors is shown in Fig. A9.

Table A3 ANOVA result of $\lambda$, $\gamma$ and $U$

| Source | F | p-value | Significant |
|---|---|---|---|
| $\lambda$ | 0.00453 | 0.9465 | No |
| $\gamma$ | 2757.814 | <0.001 | Yes |
| $U$ | 12.195 | <0.001 | Yes |

As summarized in Table A3, both the feasibility classifier guidance weight, $\gamma$, (F=12.195, p < 0.001) and resampling iteration number, $U$, (F=2757.814, p < 0.001) significantly affected the performance, whereas the resistance guidance weight, $\lambda$, showed no significant main effect (p = 0.9465).

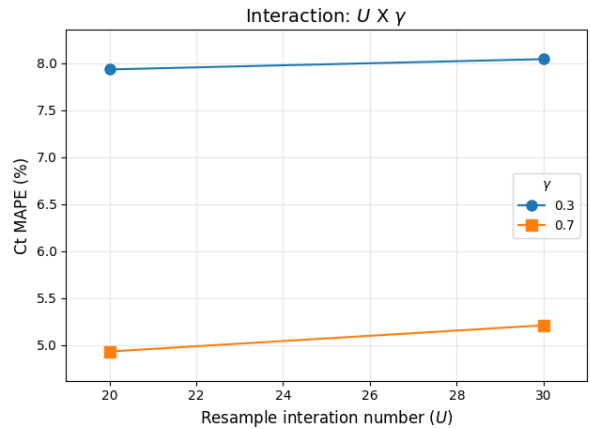

**Fig. A9 The interaction of the two most significant factors**

By observing the results of Fig. A9, the increase of $\gamma$ from 0.3 to 0.7 allows the $C_t$ MAPE to decrease from around 8% to around 5%. At the same time, the $C_t$ error shows a smaller change in the change of $U$. It indicates that the feasibility classifier guidance weight, $\gamma$, makes a more significant effect on the $C_t$ error than the resampling interaction number, $U$, and the larger $\gamma$ and lower $U$ can bring a lower $C_t$ error.